\def\year{2022}\relax
\documentclass[letterpaper]{article} 
\usepackage{aaai22}  
\usepackage{times}  
\usepackage{helvet}  
\usepackage{courier}  
\usepackage[hyphens]{url}  
\usepackage{graphicx} 
\usepackage{natbib}  
\usepackage{caption} 
\DeclareCaptionStyle{ruled}{labelfont=normalfont,labelsep=colon,strut=off} 
\usepackage{algorithm}
\usepackage{algorithmic}
\usepackage{dsfont}
\usepackage{amsmath}
\usepackage{booktabs}
\usepackage{amssymb}
\usepackage{multirow}
\usepackage{newfloat}
\usepackage{listings}
\floatstyle{ruled}
\newfloat{listing}{tb}{lst}{}
\floatname{listing}{Listing}

\newcommand{\smallsim}{\smallsym{\mathrel}{\sim}}

\makeatletter
\newcommand{\smallsym}[2]{#1{\mathpalette\make@small@sym{#2}}}
\newcommand{\make@small@sym}[2]{%
  \vcenter{\hbox{$\m@th\downgrade@style#1#2$}}%
}
\newcommand{\downgrade@style}[1]{%
  \ifx#1\displaystyle\scriptstyle\else
    \ifx#1\textstyle\scriptstyle\else
      \scriptscriptstyle
  \fi\fi
}
\makeatother

\title{Classifier Data Quality -- A Geometric Complexity Based Method for Automated Baseline And Insights Generation}
\author{
    George Kour\textsuperscript{\rm 1,2},  Marcel Zalmanovici\textsuperscript{\rm 1}, Orna Raz \textsuperscript{\rm 1}, Samuel Ackerman\textsuperscript{\rm 1}, Ateret Anaby-Tavor \textsuperscript{\rm 1}\\
    }
\affiliations{
    \textsuperscript{\rm 1}IBM Research, Haifa\\
    \textsuperscript{\rm 2}Sagol Department of Neurobiology, University of Haifa

}

\begin{document}

\maketitle

\begin{abstract}
Testing Machine Learning (ML) models and AI-Infused Applications (AIIAs), or systems that contain ML models, is highly challenging. In addition to the challenges of testing classical software, it is acceptable and expected that statistical ML models sometimes output incorrect results.  
A major challenge is to determine when the level of incorrectness, e.g., model accuracy or F1 score for classifiers, is acceptable and when it is not. In addition to business requirements that should provide a threshold, it is a best practice to require any proposed ML solution to out-perform simple baseline models, such as a decision tree.

We have developed complexity measures, which quantify how difficult given observations are to assign to their true class label; these measures can then be used to automatically determine a baseline performance threshold.
These measures are superior to the best practice baseline in that, for a linear computation cost, they also
quantify each observation' classification complexity in an explainable form, regardless of the  classifier model used.
Our experiments with both numeric synthetic data and real natural language chatbot data demonstrate that the complexity measures effectively highlight data regions and observations that are likely to be misclassfied. 

\end{abstract}

\section{Introduction} \label{sec:intro}
Testing AIIAs is highly challenging. 
The characteristics of the data used for training an ML model are key to the quality of the resulting ML model. 
The geometric foundation underlying many classifiers is to assign an observation to the closest class.
However, different classifiers may use different geometrical properties to capture the class geometry as well as may employ different distance functions to capture the closeness between an observation and a class geometry.
Here, we present a complexity heuristic based on this intuition that attempts to quantify the difficulty of assigning a given observation to its true class label by calculating the relative closeness of the observation to its true class vs. its closeness to other classes. 

The complexity measure provides insights for improving the system design, for improving labeling, or for directing the gathering or generation of more data.  It can further be used to set an expected performance level, which we call a `baseline', for a simple distance-based discriminator on the entire dataset. Section~\ref{sec:methodology} provides more details. 
Our complexity measures have the following properties:
\begin{itemize}
    \item They are computationally efficient and feasible to calculate quickly.
    \item They provide information both about single observations and about sets of observations, such as those that have the same label or class.
    \item They are general and are applicable to a wide variety of discrimination algorithms as they share similar underlying geometrical properties, stemming from Gaussian Discriminant Analysis, GDA, \cite{QDA}.
    \item They are explainable in that the geometry provides a clear reason for indicating an observation or a set of observations as complex.
\end{itemize}
Our method overcomes some of the weaknesses of the best practice baseline: it uses all the data and does not necessitate dividing the data into train and validation sets --- this is especially important when the data is small; it highlights difficult to classify observations at the same low calculation cost --- compared to manual debugging and very costly methods for scoring individual observations such as leave-one-out and Shapley values; it provides a graphical explanation of the scores (see examples in Section~\ref{sec:results}). 

We assess the properties and effectiveness of our complexity measures over numeric synthetic data. 
Then we demonstrate the effectiveness of our measures over real natural language data. We focus on the especially challenging domain of chatbots. Chatbots are a prominent example of an AIIA. Chatbots allow users to interact with the business through a natural language interface and are becoming a key channel for customer engagement. For many customers, the chatbot provides their first interaction with the business (e.g. for support purposes). Therefore, it is important for a chatbot to be of good quality from day one. 
Chatbot technology usually comprises two basic components:
\begin{enumerate}
    \item A machine learning (ML) natural language processing (NLP) based intent classifier that can process what the user is saying, and 
    \item A conversation flow orchestrator that incorporates domain knowledge and is driven by the business actions and potentially content extracted from past human-to-human dialogs and company documents.
\end{enumerate}
We utilize our complexity measures to assess the quality of the intent classifier data and to set a baseline for any intent classifier trained with that data. 

As part of our methodology we automatically find the interesting measures and ranges of complexity. These are the ranges that are more likely to result in misclassification. For that purpose we utilize our IBM FreaAI technology \cite{FreaAI1,FreaAI2}.
Automatically highlighting these ranges assists in fault identification as well as in selecting additional data for training or for testing. This data may be automatically generated, e.g., through the LAMBADA technology \cite{lambada}.

 The rest of the paper is organized as follows: Section~\ref{sec:methodology} introduces our complexity measures and their usage in automatically generating a baseline and indicating complex observations that are likely to challenge many discrimination algorithms. Section~\ref{sec:results} demonstrates how we utilize these measures for setting a baseline by identifying those observations that are likely to be misclassified. Section~\ref{sec:related} summarizes related work and Section~\ref{sec:discussion} concludes and discusses limitations of our approach. 

\section{Methodology} 
\label{sec:methodology}
In  this  section  we  describe  our  framework  for  estimating the  complexity---in terms of difficulty of assignment to its true label---of a sample in  a  labeled  data  set  or corpus, in the case of text data.
We also provide a methodology for using the resulting complexity measures when automatically setting a baseline for many common discrimination algorithms as well as providing additional insights about observations and their complexity. Complex to classify observation are highlighted in an explainable way. These may be observations that are part of the data used to compute the complexity measures, such as those in the train set, or observations yet unseen, such as those in the test set or those observations that came at runtime after deployment (e.g., user utterances after the chatbot is deployed
).

Given a labeled corpus, we first employ a text embedding model $e_\theta$ on each of the samples to obtain a dataset $\mathcal{D}=\{(x_i,y_i)\}_{i=1}^N$ where $x_i=e_\theta(s_i)$ and $x_i \in \mathcal{X} \subseteq \mathds{R}^d$ is a $d$-dimensional dense vector representation of the textual sample $s_i$, and $y_i\in \mathcal{Y}$ is its corresponding label.
In this work, we employ the \textsc{sentence-transformers} \cite{reimers-2019-sentence-bert} python package and the attention-based pre-trained language model called \textsc{paraphrase-MiniLM-L6-v2} \cite{reimers-2019-sentence-bert} which maps textual samples to $\mathds{R}^{384}$.

Our goal is to define a simple yet effective geometric-based measure $h:\mathcal{X}\times\mathcal{Y} \rightarrow \mathds{R}_{\geq 0}$ which evaluates the incompatibility of a given sample $(x,y)$ to the underlying distribution and thus estimating how difficult it would be for a discriminator to correctly predict its label.
In essence, the idea behind our complexity measure is to  calculate the relative geometrical closeness of the sample to its own class versus its closeness to other classes in the dataset.
The relevant geometric properties considered in the calculation are determined by the employed distance measure $\delta(x,B)$ which should capture the perceptual distance between the sample $x$ and the geometry of the class samples in $B$.
Thus, we define the complexity heuristic as follows:
\begin{equation}
\begin{split}
    h(x,y) &= -\log\left[\frac{e^{-\delta(x,\mathcal{C}(y))}}{\sum_{c\in \mathcal{Y}} e^{-\delta(x,\mathcal{C}(c))}}\right]\\
    &= \delta(x,y)+\log\left[\sum_{c\in \mathcal{Y}} e^{-\delta(x,\mathcal{C}(c))} \right]
\end{split}
\end{equation}
where $\mathcal{C}(c)$ represents the set of the samples in $\mathcal{D}$ that belong to class $c$ (namely, $\mathcal{C}(c)=\{x|(x,c)\in \mathcal{D}\}$).



We compare multiple distance functions which capture different geometric properties: Euclidean ($E$), Cosine similarity ($S$), and Mahalanobis ($M$) \cite{mclachlan1999mahalanobis}, defined respectively in Equations \ref{eq:euclidean}, \ref{eq:cosine} and \ref{eq:mahalanobis} below.
Note that while the Euclidean and the cosine metrics consider only the classes' centroids, the Mahalanobis distance captures the dispersion of the classes along each of the embedding dimensions, and calculates the ``effective" distance of the sample from the class geometry.

\begin{equation}
    \delta_E(a,B)=||a-\hat{\mu}_B||_2
\label{eq:euclidean}
\end{equation}

\begin{equation}
    \delta_S(a,B)=\frac{a \cdot \hat{\mu}_B}{\|a\|\|\hat{\mu}_B \|}
\label{eq:cosine}
\end{equation}

\begin{equation}
    \delta_M(a,B)=\sqrt{(a-\hat{\mu}_B)^T \hat{\Sigma}_B^{-1} (a-\hat{\mu}_B)}
\label{eq:mahalanobis}
\end{equation}
where $B$ is a class of samples, and $\hat{\mu}_B$ and $\hat{\Sigma}_B$ are the set centroid and empirical covariance matrix, calculated as follows:
\begin{equation}
\hat{\mu}_B= \frac{1}{|B|}\sum_{b\in B} b
\end{equation}

\begin{equation}
\hat{\Sigma}_B=\frac{1}{|B|}\sum_{b \in B} (b-\hat{\mu}_B)(b-\hat{\mu}_B)^T
\label{eq:covariance_estimation}
\end{equation}

Using Mahalanobis distance in the complexity measure is equivalent to calculating negative log-likelihood loss of a generative classification model which fits a class-conditioned Gaussian distribution to every class of the data.
In \cite{lee2018simple} a similar method is used to calculate confidence scores for out-of-distribution (OOD) detection in test samples, however, in their case the confidence function $M(x)$ is independent of the given sample label and thus for a given feature vector $x$, they defined the confidence score as follows:

\begin{equation}
    M(x)=\max_c \{- \delta_M(x,\mathcal{C}(c))\}
\end{equation}

Despite being an asymptotically unbiased estimator of the covariance matrix, the Maximum Likelihood Estimator (Equation \ref{eq:covariance_estimation}) is imperfect and the precision matrix obtained from its inversion may be inaccurate and sometimes even unobtainable due to numerical reasons.
Estimating the covariance matrix is even harder in cases where the number of samples is smaller than the number of features (i.e., $n<p$). 
Thus, in relatively small datasets, the empirical covariance matrix is commonly underestimated.
To mitigate the estimation error of the covariance matrix in high-dimensional feature spaces shrinkage-based covariance estimation methods, such as the Ledoit-Wolf shrinkage approach \cite{ledoit2004well}, are commonly employed.

We found that using the estimated Pearson correlation coefficients matrix \cite{galton1877typical} instead of the covariance matrix in the Mahanalabis distance for calculating the complexity yields better results, in the sense that it better correlates with the error concentration of a prototypical classifier (more information is provided in Section \ref{sec:results}).
In the following, using the correlation matrix is denoted by $\delta_{\hat{M}}$.
We hypothesise that despite the lose of information about the manifold radii, using correlation coefficients matrix is the better option in this case as it is less affected by the number of samples.

The complexity measure may be employed on both the train and test sets.
While its goal in both cases is to identify difficult to classify samples, its practical interpretation can be different depending on the usage setup.
For instance, when used on the train set it may be used to remove training samples which may be mis-labeled or spot samples with high information gain; however, when used on the test set, high complexity samples can be viewed as ambiguous or out-of-domain.
Note that when the complexity measure is used to rank the train samples, their classification difficulty may be underestimated because the estimated geometry of the classes is affected by the samples to be ranked. 
However, when having enough samples, the marginal effect of each single sample on the estimated geometry is small.





 \subsection{Setting a baseline} 
 \label{sub:baseline}
 In our framework, given a dataset $\mathcal{D}$ and a distance function $\delta(x,B)$, where x is a sample in the $d$-dimensional space and $B$ is a set of samples, the baseline classifier is  defined as follows:

 \begin{equation}
     f(x)=\textrm{argmin}_{c \in \mathcal{Y}} \delta(x,C(c))
 \end{equation}
 
Note that the complexity measure is defined as the negative log-likelihood of $f(x)$.
Using $\delta_E$ or $\delta_S$ distance functions, the classifier becomes  a Nearest-Centroid classifier in terms of amplitude or angle, respectively.  Observations that are more difficult to classify by class-centroid distance will receive higher complexity scores under our heuristic.

However, when the Mahalanobis distance is used, i.e., this classifier becomes a Gussian Discriminant Analysis (GDA) classifier, and when the Pearson correlation coefficient is used instead of the covariance matrix then the classifier is reduced to Quadratic Discriminante Analysis classifier (QDA).
These classifiers are attractive because they have been proven to work, have no hyperparameters to tune, and have closed-form solutions that can be easily computed.


\section{Experiments and results} \label{sec:results}
Evaluating the complexity measure involves assessing its correlation with both the errors and confidence of a somehow prototypical classifier.
Namely, high complexity values are expected for samples on which the classifier tends to err or for samples that may be classified correctly by the classifier but with low confidence.
Moreover, when a trained classifier errs with high confidence, the complexity measure may correctly indicate high complexity. 

Our experimental results indicate that the complexity measures are similar to or even out-perform a trained classifier confidence measure.
This is especially encouraging as it means that the complexity measures can reasonably estimate the difficulty a classifier would have to correctly classify samples, even prior to any training.


We experiment with both synthetic numeric data as described in Section \ref{subsec:synthetic} and with real natural language chatbot data as described in Section \ref{sub:text}.
We utilize FreaAI for automatically finding the most correlating features and ranges with high misclassification concentration, which we refer to as \emph{Slices}. 
For each slice computed by FreaAI, a relative rank is calculated, which can be intuitively interpreted as follows: the higher the ranking, the higher the error concentration of the classifier on that slice compared to overall performance.
In addition to the performance of the classifier in the slice, the ranking also considers the number of samples in the dataset that fall into the slice. 

\subsection{Synthetic data} \label{subsec:synthetic}

Note that for different needs, the complexity measure can be employed on either the train samples or on other samples not used for training.
In the train use-case case, the geometries and classifier are learned approximated using the train set, and the complexity is calculated on the same train samples.
However, in the test use-case, the training and the approximation of the geometries is performed using the train set, but the complexity is calculated on the new samples.

To demonstrate the complexity measure potency in both the train and test use-cases, our experiments include calculating the complexity measure on the train set of synthetic $2$-dimensional data, and on the test set of a real-life sentence utterance classification (SUC) datasets. 

In the first set of experiments, we generated synthetic samples for three classes $c_0, c_1, c_2$ by drawing random samples from three bivariate normal distributions. 
The distributions are specified by their mean and covariance matrix.
Nevertheless, the classes geometries partially overlap to simulate ambiguous samples.
Second, the complexity measure is calculated for each sample and SVM classifier \cite{SVC} is trained to predict the sample classes.
Then, the classifier is used to predict the samples labels and the prediction is analyzed using FreaAI technology to find and rank slices with high error concentrations. 
Over-fitting of the classifier is avoided by applying a relatively simple model, and thus, we expect most errors in predicting the labels to be in the overlap.
The advantage of using synthetic datasets is that it makes the analysis and visual validation easier. 
The main purpose is to make sure that the complexity highlights areas where it is more likely to encounter errors and that the FreaAI technology identifies the expected complexity ranges as problematic.
The data records sent to FreaAI contain for each sample its coordinates $x_1$, $x_2$, the true label $y$, the label predicted by the trained classifier $\hat{y}$ and its prediction confidence as well as the different complexity measures proposed in Section \ref{sec:methodology}.



\begin{figure}
    \centering
    \includegraphics[scale=0.3]{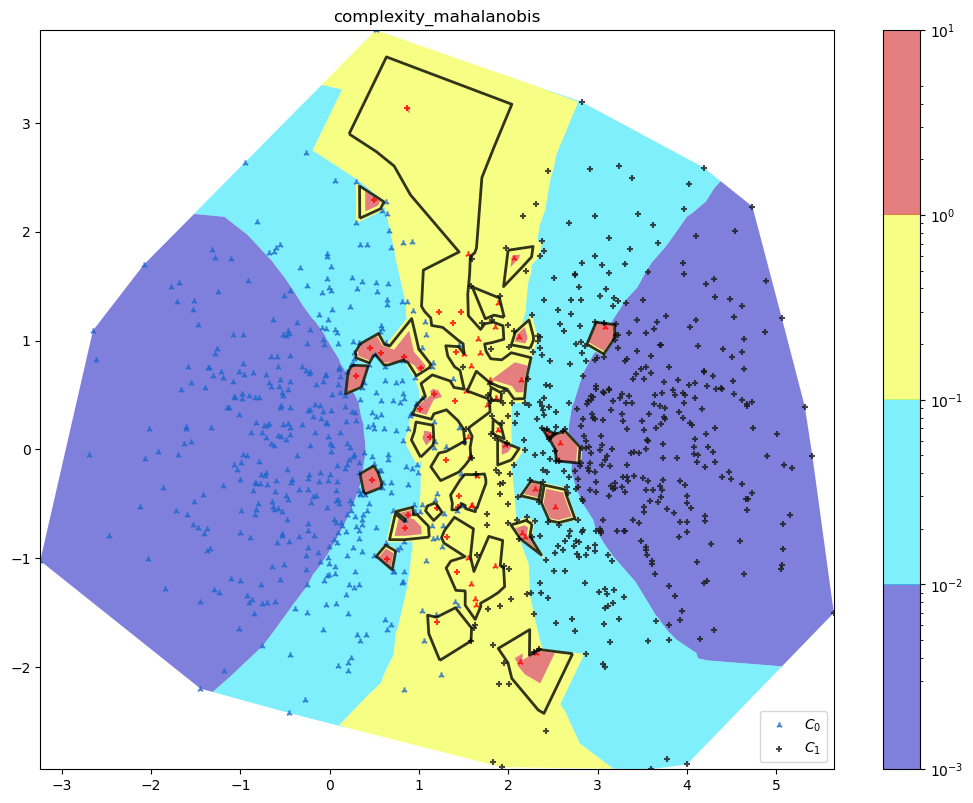}
    \caption{\label{fig:synth_2_classes} Synthetic data drawn from two Gaussians of equal size; background the complexity calculated via the Mahalanobis method. In all Figures the background shows the complexity areas that  were  interpolated  by  the  samples  in  the  dataset.  The color legend goes from most complex at the top to least complex at the bottom of the legend. The samples and their class label are represented by colored points. The areas bounded by the black contour represent slices found by FreaAI and contain high concentration of errors.}
\end{figure}

In the first and simplest experiment we use only two classes of the same size that overlap as seen in Figure \ref{fig:synth_2_classes}. 
In this case we expect most errors to be located in the overlap as well as for those samples to be scored with higher complex. 
The generated data has 500 samples in each class and 63 errors in total (accuracy 0.94).
The FreaAI slices on the Mahalanobis and Euclidean complexity confirm that this is where the errors are found as can be seen in Figure \ref{fig:synth_2_classes}. 

The background of all figures with a complexity measure is similar to Figure \ref{fig:synth_2_classes} and shows the complexity areas that were interpolated by the samples in the dataset. The color legend goes from most complex at the top to least complex at the bottom of the legend. 
The samples and their class label are represented by colored points.
The areas bounded by the black contour represent slices found by FreaAI and contain high concentration of errors.  

All experiments similarly capture the higher complexity ranges that correlate best with misclassifications. Figure \ref{fig:synth_2_classes} and its corresponding Table  \ref{fig:synth_2_classes} mark, for example, a range where 65 records appear which contains all 63 errors in this data. The cosine complexity measure marked a much larger range and still didn't cover most errors. The model confidence in this case shows that all errors are when the confidence is less than 0.96 which does not help much.
We show also a few interesting ranges found by FreaAI on $x_1$, $x_2$ or both where relatively many errors are found. Not surprisingly these areas are all in the overlap between the classes.


\begin{table}[htbp]
  \small
  \centering
    \begin{tabular}{p{1.7cm}p{1.7cm}p{1cm}p{8mm}p{8mm}}
    \toprule
    \textbf{Feature(s)}  & \textbf{Slice} & \textbf{$acc_{SVC}$}  & \textbf{Size} & \textbf{Rank}\\
    \midrule
    \textbf{compl\_mah} &  0.46 $\smallsim$ 5.40                 & 0.05 & 65  & 0.99 \\
    \textbf{compl\_euc} &  0.46 $\smallsim$ 5.40                 & 0.05 & 65  & 9.98 \\
    $x_1$               &  0.30 $\smallsim$ 2.57                 & 0.81 & 330 & 0.58 \\
    \textbf{confidence} &  0.50 $\smallsim$ 0.96                 & 0.80 & 309 & 0.56 \\
    \textbf{compl\_cos} &  0.34 $\smallsim$ 1.61                 & 0.84 & 168 & 0.55 \\
    $x_1$, $x_2$     &  1.52 $\smallsim$ 1.69\newline-1.43 $\smallsim$ 1.80 & 0.26 & 19 & 0.54 \\
    $x_1$, $x_2$     &  1.20 $\smallsim$ 1.44\newline-0.54 $\smallsim$ 1.26 & 0.27 & 11 & 0.35 \\
    \textbf{compl\_cos} &  2.38 $\smallsim$ 2.41                 & 0.58 & 12 & 0.15 \\
    $x_2$               &  0.85 $\smallsim$ 1.89                 & 0.71 & 17 & 0.13 \\
    $x_2$               & -0.54 $\smallsim$ -0.51                & 0.54 & 11 & 0.10 \\
    \bottomrule
    \end{tabular}%
  \label{tab:2classes}%
  \captionsetup{width=.95\columnwidth}  
  \caption{Slices for two Gaussians of equal size. These slices are depicted in Figure \ref{fig:synth_2_classes} as areas bounded by black contour. }
\end{table}%

\begin{figure}
    \centering
    \includegraphics[scale=0.3]{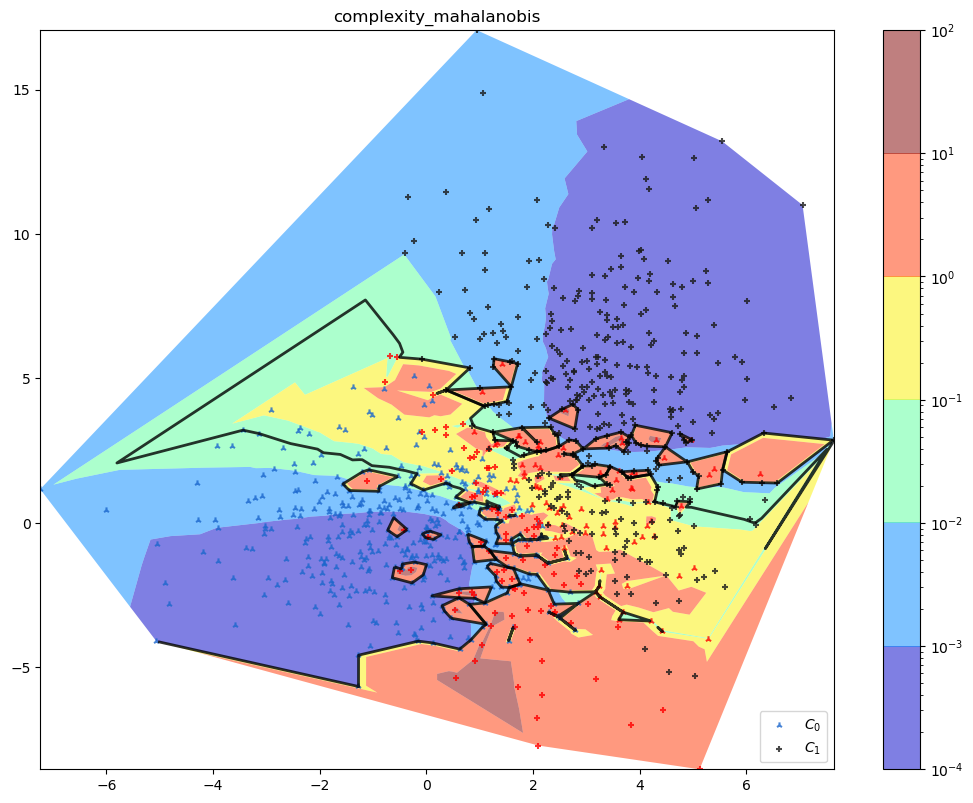}
    \caption{\label{fig:synth_2_classes_diff_mani} Similar to Figure \ref{fig:synth_2_classes} however here the shape of one class is still a circle while the other is an ellipse.}
\end{figure}

For the second experiment we turned one of the classes into an ellipse as depicted in Figure \ref{fig:synth_2_classes_diff_mani}.
This is to verify that the complexity works as expected when the distance on each dimension varies.
As can be seen, the higher complexity is marked in the main intersection of the Gaussians and also in the bottom-right and top-left sides where the narrower ellipse points are drawn. This data has 181 errors (accuracy 0.82). For example the Mahanalobis complexity range identified by FreaAI contains 141 of those errors, versus 371 found in the Euclidean complexity and 76 in the "best" confidence range. 

\begin{table}[htbp]
  \small
  \centering
    \begin{tabular}{p{1.7cm}p{1.7cm}p{1cm}p{8mm}p{8mm}}
    \toprule
    \textbf{Feature(s)}  & \textbf{Slice} & \textbf{$acc_{SVC}$}  & \textbf{Size} & \textbf{Rank}\\
    \midrule
    \textbf{compl\_mah}  &  0.04 $\smallsim$ 17.87                  & 0.45 & 320 & 0.51 \\
    \textbf{compl\_euc}  &  0.03 $\smallsim$ 18.63                  & 0.49 & 351 & 0.48\\
    $x_1$, $x_2$         &  0.34 $\smallsim$ 5.30\newline-8.53 $\smallsim$ -2.88 & 0.36 & 47  & 0.47 \\
    $x_1$, $x_2$         &  0.82 $\smallsim$ 4.27\newline-2.81 $\smallsim$ 3.14  & 0.63 & 323 & 0.45 \\
    \textbf{compl\_cos}  &  1.32 $\smallsim$ 2.83                   & 0.55 & 182 & 0.42 \\
    \textbf{confidence}  &  0.50 $\smallsim$ 0.63                   & 0.50 & 151 & 0.40 \\
    $x_1$                &  1.80 $\smallsim$ 2.60                   & 0.61 & 137 & 0.38 \\
    $x_2$                &  8.53 $\smallsim$ 3.02                    & 0.55 & 64  & 0.32 \\
    \bottomrule
    \end{tabular}%
  \label{tab:2classes_unequal}%
  \captionsetup{width=.95\columnwidth}  
  \caption{Slices for two Gaussians of unequal size as shown in Figure \ref{fig:synth_2_classes_diff_mani}. }
\end{table}%

\begin{figure}
    \centering
    \includegraphics[scale=0.3]{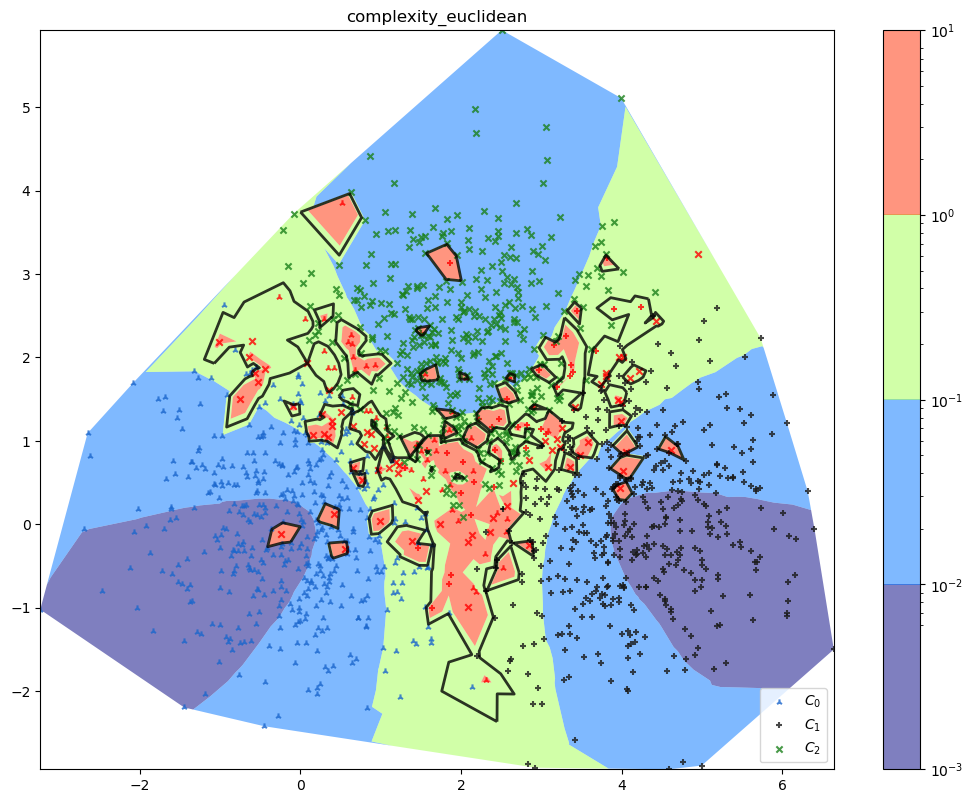}
    \caption{\label{fig:synth_3_classes_single_overlap_mah} Synthetic data with three Gaussians overlapping in the same area; background shows the Euclidean complexity.}
\end{figure}

\begin{figure}
    \centering
    \includegraphics[scale=0.3]{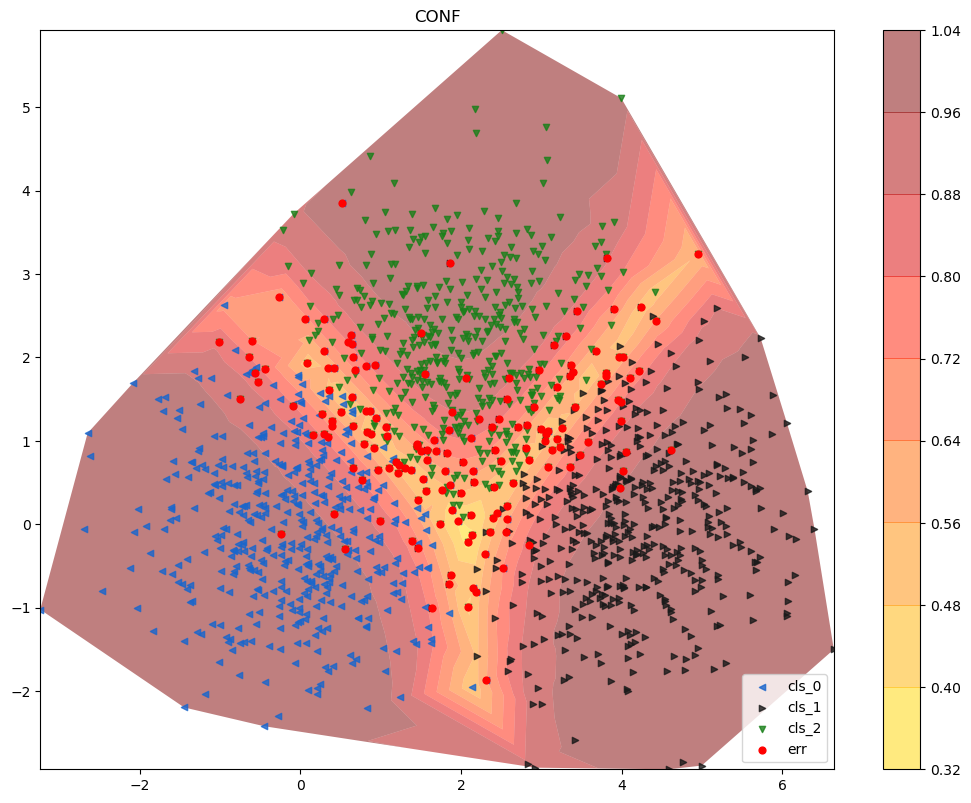}
    \caption{\label{fig:synth_3_classes_single_overlap_conf} The same data as in Figure \ref{fig:synth_3_classes_single_overlap_mah} with the background showing the model confidence.}
\end{figure}

For the third experiment we generated three same-sized Gaussians that overlap in roughly the same area.
This data contains 160 mistakes (accuracy 0.89).
Figure \ref{fig:synth_3_classes_single_overlap_mah} shows the Euclidean complexity as the background for the data. This is almost identical to the heat map generated by the Mahalanobis complexity and quite similar to the model confidence heat map shown in Figure \ref{fig:synth_3_classes_single_overlap_conf}.

\begin{table}[htbp]
  \small
  \centering
    \begin{tabular}{p{1.7cm}p{1.7cm}p{1cm}p{8mm}p{8mm}}
    \toprule
    \textbf{Feature(s)}  & \textbf{Slice} & \textbf{$acc_{SVC}$}  & \textbf{Size} & \textbf{Rank}\\
    \midrule
    \textbf{compl\_euc} &  0.54 $\smallsim$ 8.93                & 0.16 & 187 & 0.93 \\
    \textbf{compl\_mah} &  0.54 $\smallsim$ 8.98                & 0.16 & 187 & 0.92 \\
    \textbf{compl\_cos} &  0.93 $\smallsim$ 3.58                & 0.76 & 485 & 0.49 \\
    \textbf{confidence} &  0.37 $\smallsim$ 0.86                & 0.66 & 426 & 0.48 \\
    $x_1$, $x_2$        & -0.98 $\smallsim$ 2.91\newline0.62 $\smallsim$ 1.42  & 0.73 & 185 & 0.45\\
    $x_2$               &  0.87 $\smallsim$ 1.36                & 0.74 & 173 & 0.44 \\
    $x_1$, $x_2$        &  2.98 $\smallsim$ 4.22\newline0.64 $\smallsim$ 2.27  & 0.65 & 95 & 0.40\\
    $x_1$               &  1.45 $\smallsim$ 1.58                & 0.76 & 46 & 0.32 \\
    \bottomrule
    \end{tabular}%
  \label{tab:3classes_one_overlap}%
  \captionsetup{width=.95\columnwidth}  
  \caption{Slices for three Gaussians of equal size. (Some) of the results of the data shown in Figure \ref{fig:synth_3_classes_single_overlap_mah} and \ref{fig:synth_3_classes_single_overlap_conf}.}
\end{table}%

\begin{figure}
    \centering
    \includegraphics[scale=0.3]{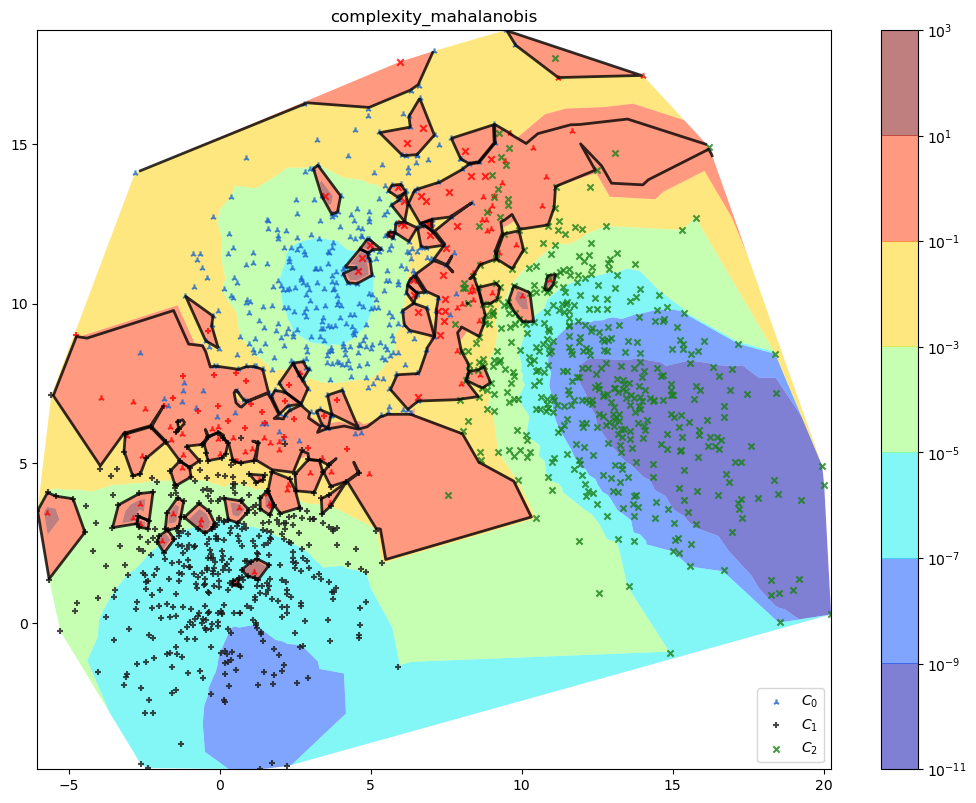}
    \caption{\label{fig:synth_3_classes_two_overlaps_mah} Three Gaussians with two distinct overlapping areas; background is Mahalanobis complexity.}
\end{figure}

In the fourth experiment we used three classes: two elliptic Gaussians and one round. To make it more interesting we also made sure that $x_1$ overlapped with $x_2$ in one area and with $x_3$ in another. The result can be seen in Figures \ref{fig:synth_3_classes_two_overlaps_mah}-- 
\ref{fig:synth_3_classes_two_overlaps_conf}.


\begin{figure}
    \centering
    \includegraphics[scale=0.3]{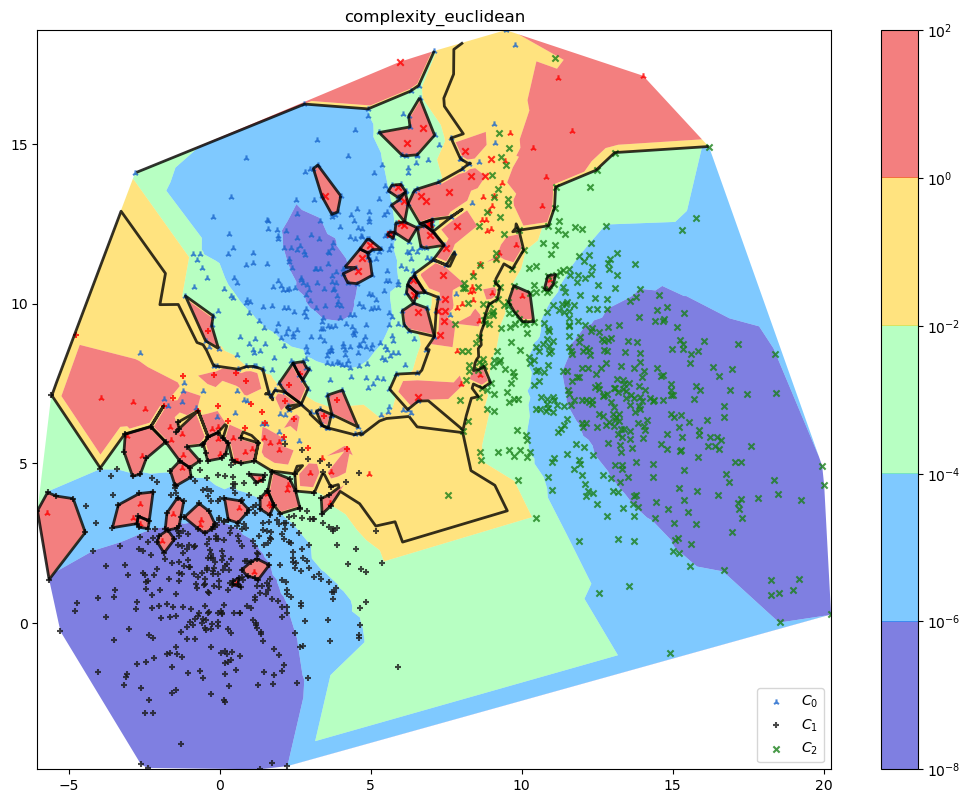}
    \caption{\label{fig:synth_3_classes_two_overlaps_euc} Three Gaussians with two distinct overlapping areas; background is Euclidean complexity.}
\end{figure}


\begin{table}[htbp]
  \small
  \centering
    \begin{tabular}{p{1.6cm}p{1.9cm}p{1.2cm}p{8mm}p{8mm}}
    \toprule
    \textbf{Feature(s)}  & \textbf{Slice} & \textbf{$acc_{SVC}$}  & \textbf{Size} & \textbf{Rank}\\
    \midrule
    \textbf{compl\_mah} &  0.14 $\smallsim$ 62.43               & 0.25 & 178 & 0.90 \\
    \textbf{compl\_euc} &  0.07 $\smallsim$ 71.13               & 0.34 & 209 & 0.79 \\
    \textbf{confidence} &  0.47 $\smallsim$ 0.83                & 0.67 & 310 & 0.52 \\
    $x_1$, $x_2$        & -6.04 $\smallsim$ 8.62\newline3.06 $\smallsim$ 7.81  & 0.76 & 279 & 0.51\\
    \textbf{compl\_cos} &  1.08 $\smallsim$ 1.60                & 0.66 & 255 & 0.50 \\
    $x_1$, $x_2$        & 5.90 $\smallsim$ 10.95\newline7.83 $\smallsim$ 18.56 & 0.76 & 243 & 0.49 \\
    $x_1$               &  5.90 $\smallsim$ 9.45                & 0.74 & 207 & 0.48 \\
    $x_2$               & 12.14 $\smallsim$ 18.56               & 0.78 & 165 & 0.45 \\
    \bottomrule
    \end{tabular}%
  \label{tab:3classes_two_overlaps}%
  \captionsetup{width=.95\columnwidth}  
  \caption{\label{fig:synth_3_classes_two_overlaps_conf} Slices for three Gaussians with multiple overlapping areas. Three Gaussians with two distinct overlapping areas; background is the model confidence. Top results of the data shown in Figures \ref{fig:synth_3_classes_two_overlaps_mah}, \ref{fig:synth_3_classes_two_overlaps_euc} and \ref{fig:synth_3_classes_two_overlaps_conf} }
\end{table}%


\subsection{Real textual data} \label{sub:text} 

The experimental methodology performed in the real-life textual datasets is similar to the methodology used in the synthetic data except that here, the complexity is calculated on the test set while the class geometries ere estimated on the train set.
Also, as expected, the classifier was trained on the train set and evaluated on the test set.
The split between train and test samples was performed randomly, maintaining 80\%-20\% train-test ratio.

We also experimented with following four textual data-sets:

\begin{itemize}
    \item ATIS (\citealt{ATIS}) is a standard benchmark data-set widely used as an intent classification. The test part has 800 records in 8 classes. The model we trained on it had 0.97 accuracy, meaning only 22 errors.
    
    \item WEBRR (Web Answer Passages; \citealt{webap_data}) is a TREC GOV2 collection with 80 questions that serve as the labels and their answers, which serve as the samples; the answers were ranked and validated by users.
    
    \item NEWSGROUPS (\citealt{misra2021sculpting}; \citealt{misra2018news}) is a collection of newsgroup documents. We used the 10 classes with the most documents. The model performed rather poorly with an accuracy of 0.57 (1060 errors).
    
    \item HWU64 \cite{liu2019benchmarking} is Part of the DialoGLUE Benchmark containing popular personal assistant queries. The test set has 4108 records. The trained model achieve 0.80 accuracy (814 errors).
\end{itemize}
Table \ref{tab:ds_stat} provides a basic statistical overview on these dataset.
Note that the normalized entropy of the samples' distribution over the classes is used to estimate the degree of imbalance in the dataset. 
We formally define it as 
\begin{equation}
    \mathcal{E}(\mathcal{D})=\frac{-\sum_{c\in\mathcal{Y}}{ \frac{|\mathcal{C}(c)|}{|\mathcal{D}|} \cdot \log{\frac{|\mathcal{C}(c)|}{|\mathcal{D}|}}}}{\log{(|\mathcal{Y}|)}}
\label{eq:entropy}
\end{equation}

\begin{table}[]
\begin{tabular}{lccccc}
\toprule
Dataset ($\mathcal{D}$)     & $|\mathcal{Y}|$ & $|\mathcal{D}|$ &  $\mathcal{E}(\mathcal{D})$ & MCS  & $acc_{B}$\\ 
\midrule
ATIS            &    8      &  4834      & 0.46  & 152  & 0.97  \\
HWU64            & 60         & 20534      & 0.91    & 229 & 0.97  \\
WEBRR            & 80         & 3298       & 0.88    & 25 & 0.57  \\
NEWSGROUPS        & 41         & 50555      & 0.99    & 1250&  0.8 \\
\bottomrule
\end{tabular}
\caption{Dataset distribution statistics. In accordance with the notation in Section \ref{sec:methodology}, $|\mathcal{Y}|$ denotes the number of classes, $|\mathcal{D}|$ denotes the total number of samples. $\mathcal{E}(\mathcal{D})$ is the the normalized entropy as defined in Equation \ref{eq:entropy}, MCS is the median class size and $acc_{B}$ denotes baseline accuracy according to the baseline classifier defined in Section \ref{sub:baseline}}.
\label{tab:ds_stat}
\end{table}

With the ATIS data-set the model confidence was by far superior in its estimate of how good it would do on the records.
All other methods had similar results, with the Mahalanobis method being considered somewhat better by FreaAI when considering the trade-off between size of slice and its performance. 

Table \ref{tab:sentence_examples} shows examples of high complexity and low complexity utterances from these data-sets. It can be easily seen that complex utterances tend to be longer and include words which by themselves could belong to either class. The complex utterances also tend to be more ambiguous in their wording.

\begin{table*}
  \small
  \centering
    \begin{tabular}{p{8mm}p{8mm}p{8mm}p{8mm}p{11cm}}
    \toprule
    \textbf{Dataset}  & \textbf{$h(x,y)$} & \textbf{$y$} & \textbf{$\hat{y}$} & \textbf{Utterance} \\
    \midrule
    \multirow{4}{*}{ATIS} & 3.68 & ftime  & flight   & what are the departure times from detroit to westchester county \\
                          & 3.15 & flight & aircraft  & show me the connecting flights between boston and denver and the types of aircraft used \\
                          & 0.05 & abbrv & abbrv & what does fare code qo mean \\
    \midrule
    \multirow{4}{*}{News} & 13.47 & taste & good & These Heroic Food Trucks Are Coming To The Rescue Of California Fire Victims \\
                          & 11.31 & healthy & crime & Golden Gate Bridge Finally Getting A Suicide Barrier \\
                          & 0.11  & crime & crime & Two Police Officers Killed In Palm Springs, California Shooting \\
    \midrule
    \multirow{4}{*}{Web RR} & 23.62 & rule & relation & If there are entrances in more than two directions, the union should be to the East. \\
                            & 14.25 & cult & church & In 1989,two members of a church, described by ATP as a doomsday religious cult... \\
                            & 0.001 & orange & orange & Tangerines \\
    \bottomrule
    \end{tabular}%
  \captionsetup{width=.95\textwidth}  
  \caption{Utterance examples: Examples of complex and simple utterances from the data-sets. Column \textbf{$h(x,y)$} is the complexity measure based on the Mahalanobis distance function; \textbf{$y$} and \textbf{$\hat{y}$} are the utterance class and predicted class. Class name are shortened to fit the limited table space, especially for Web RR where these are phrases.}
  \label{tab:sentence_examples}%
\end{table*}%

\begin{table}
  \small
  \centering
    \begin{tabular}{p{9mm}@{\hspace{3mm}}p{5mm}@{\hspace{2mm}}p{1.1cm}p{1.75cm}@{\hspace{3mm}}p{5mm}@{\hspace{3mm}}p{5mm}@{\hspace{3mm}}p{5mm}}
    \toprule
    \textbf{Dataset}  & \textbf{$acc$} & \textbf{Metric} & \textbf{Slice} & \textbf{Slice $acc$}  & \textbf{Size} & \textbf{Rank} \\
    \midrule
    \multirow{4}{*}{ATIS} & \multirow{4}{*}{0.97} & \textbf{\textit{confidence}}  & 0.03 $\smallsim$ 0.47 & \textbf{\textit{0.15}}  & \textbf{\textit{26}}  & \textbf{\textit{0.86}} \\
                          &                       & compl\_mah           & 0.31 $\smallsim$ 3.69 & 0.88  & 124 & 0.52 \\
                          &                       & compl\_euc           & 2.23 $\smallsim$ 4.54 & 0.82  & 93  & 0.49 \\
                          &                       & compl\_cos           & 3.81 $\smallsim$ 4.19 & 0.79  & 72  & 0.46 \\
    \midrule
    \multirow{4}{*}{Web RR} & \multirow{4}{*}{0.97} & confidence           & 0.02 $\smallsim$ 0.17    & 0.69  & 52  & 0.89 \\
                          &                         & \textbf{\textit{compl\_mah}}  & 4.31 $\smallsim$ 54.41   & \textbf{\textit{0.89}}  & \textbf{\textit{153}} & \textbf{\textit{1.00}} \\                       
                          &                         & compl\_euc           & 7.93 $\smallsim$ 20.16   & 0.86  & 121 & 0.93 \\
                          &                         & compl\_cos           & 14.18 $\smallsim$ 17.58  & 0.85  & 112 & 0.85 \\
    \midrule
    \multirow{4}{*}{News} & \multirow{4}{*}{0.57} & confidence            & 0.00 $\smallsim$  0.47    & 0.22  & 1294  & 0.57 \\
                          &                       & compl\_mah            & 4.70 $\smallsim$ 13.47    & 0.15  & 290   & 0.76 \\
                          &                       & \textbf{\textit{compl\_euc}}   & 4.46 $\smallsim$  8.05    & \textbf{\textit{0.12}}  & \textbf{\textit{382}}   & \textbf{\textit{0.89}} \\
                          &                       & compl\_cos            & 4.93 $\smallsim$  6.08    & 0.12  & 357   & 0.85 \\
    \midrule
    \multirow{4}{*}{HWU64} & \multirow{4}{*}{0.80} & \textbf{\textit{confidence}}  & 0.00 $\smallsim$ 0.35    & \textbf{\textit{0.31}}  & \textbf{\textit{1133}} &  \textbf{\textit{0.79}} \\
                          &                        & compl\_mah           & 1.35 $\smallsim$ 119.6   & 0.59  & 1680 & 0.58 \\
                          &                        & compl\_cos           & 13.63 $\smallsim$ 17.67  & 0.57  & 1574 & 0.57 \\
                          &                        & compl\_euc           & 6.05 $\smallsim$ 29.01   & 0.55  & 1514 & 0.56 \\
    \bottomrule
    \end{tabular}%
  \label{tab:text_data_sets_results}%
  \captionsetup{width=.95\columnwidth}  
  \caption{Summary of the results for all textual data-sets. The table summarizes the results over all text data-sets. \textbf{Acc} is the accuracy of the model trained on the data-set, \textbf{Metric} is the information FreaAI used to search for slices, \textbf{Slice} is the range on the feature (except for confidence, the maximum is also the maximum in the data), \textbf{Slice Acc} is the accuracy of the records in the slice using the trained model, \textbf{Size} is the support or number of records in the slice and \textbf{Rank} is the FreaAI score for the slice (higher is more important). The metric which, according to FreaAI algorithm, provides the best information is marked in bold plus italic. Except for ATIS the complexity measures, especially the Mahalanobis one, are fairly close or better than the confidence of a \textit{trained} model which means we can estimate before training a model areas which are hard to learn.}
\end{table}%

To assess the affect of employing the correlation matrix instead of the covariance matrix when calculating the Mahalanobis distance, as suggested in Section \ref{sec:methodology}, in Table \ref{tab:normalized_mah_comparison} we report the ranking and the order that FreaAI yields for the two options for each of the datasets.
We can see that on 3 of the 4 datasets the ranking of the Mahalanbis complexity measure is higher than when using the covariance matrix. 

\begin{table}[htbp]
  \centering
\begin{tabular}{lcccc}
\toprule
Dataset ($\mathcal{D}$)     & $\delta_{M}$& $\delta_{M}$ & $\delta_{\hat{M}}$ & $\delta_{\hat{M}}$ \\
 & Rank     & Order   & Order & Rank\\ 
\midrule
ATIS             &  0.54 & 2    &  \textbf{0.88}   & 2  \\
HWU64            & 0.56  & 3    & \textbf{0.6}    & \textbf{2}  \\
WEBRR            & \textbf{0.94}  & 2    & 0.89    & \textbf{1}   \\
NEWSGROUPS       & 0.53  & 4    & \textbf{0.75}    & \textbf{3} \\
\bottomrule
\end{tabular}
\captionsetup{width=.95\columnwidth} 
\caption{Comparing FreaAI ranking score and order of the Mahalanobis-based complexity using the estimated covariance, $\delta_{M}$, versus using the Pearson correlation matrix, $\delta_{\hat{M}}$.}
\label{tab:normalized_mah_comparison}
\end{table}

\section{Related work} \label{sec:related}
The geometrical properties underlying our complexity measures are based on GDA (Gaussian discriminant analysis\cite{QDA}) and thus underlie a wide variety of discrimination algorithms.
This enables to use our measures as a black box, independent of the actual discrimination algorithm used to solve the classification problem. 
Our complexity methods are similar to those developed for the purpose of separability \cite{ThorntonSepIdx}. The motivation for the separability measures is often feature selection \cite{noteOnSep, sepFeature1, setFeature2}, dealing with the features to select to best separate the data, therefore achieving better discrimination among classes or clustering, and the evaluation of different clustering techniques \cite{sepIdx,DistSepIdx}. Our motivation is different and is focused on automatically setting a baseline for testing ML discriminators. The different motivation also entails differences in implementation. For example, our work is efficiently different, as we consider the relationship between the different classes as our basis of complexity measure. Also, our method provides a score per record  for the same computational cost, which is much more efficient than the common practice methods such as leave-one-out or Shapley values. 
\section{Conclusions and discussion}
\label{sec:discussion}
We developed complexity measures and demonstrated their usage and usefulness for automatically setting a classification baseline as they highlight observations that are likely to be misclassified regardless of the discrimination algorithm used to classify them. 
Our method provides insights that are explainable via the complexity measure geometry, and does so at a linear calculation cost. 
We demoed our measures and methodology both over synthetic numeric data where we could easily validate the results, and over real natural language chatbot data. 
We believe that our complexity measures can have additional usages, such as assessing data quality and aiding in system design.

Our method may have several shortcomings including: (1) The complexity measure is calculated based on a text embedding and thus is limited by the embedding representation power. Therefore, one should select the embedding that best captures the important semantic aspects of the task. (2) The estimation of the class geometrical properties may not be accurate in small data-sets and thus one should take into account the size of the data-set when selecting the geometrical features to consider. (3) The complexity measure may return biased results in highly unbalanced data-sets. However, this limitation may be overcame using under-sampling techniques, similar to the methods employed in the field of active-learning to obtain a representative but smaller set of samples in each class. 
Alternatively, if possible, data augmentation techniques can be employed on the smaller classes to obtain a more balanced data-set.




\small {
\bibliography{main} 

\begin{thebibliography}{21}
\providecommand{\natexlab}[1]{#1}

\bibitem[{Ackerman, Raz, and Zalmanovici(2020)}]{FreaAI2}
Ackerman, S.; Raz, O.; and Zalmanovici, M. 2020.
\newblock {FreaAI}: Automated Extraction of Data Slices to Test Machine
  Learning Models.
\newblock In Shehory, O.; Farchi, E.; and Barash, G., eds., \emph{Engineering
  Dependable and Secure Machine Learning Systems}. Springer International
  Publishing.

\bibitem[{Anaby-Tavor et~al.(2020)Anaby-Tavor, Carmeli, Goldbraich, Kantor,
  Kour, Shlomov, Tepper, and Zwerdling}]{lambada}
Anaby-Tavor, A.; Carmeli, B.; Goldbraich, E.; Kantor, A.; Kour, G.; Shlomov,
  S.; Tepper, N.; and Zwerdling, N. 2020.
\newblock Do Not Have Enough Data? Deep Learning to the Rescue!
\newblock \emph{Proceedings of the AAAI Conference on Artificial Intelligence},
  34(05): 7383--7390.

\bibitem[{Barash et~al.(2019)Barash, Farchi, Jayaraman, Raz, Tzoref-Brill, and
  Zalmanovici}]{FreaAI1}
Barash, G.; Farchi, E.; Jayaraman, I.; Raz, O.; Tzoref-Brill, R.; and
  Zalmanovici, M. 2019.
\newblock Bridging the Gap between ML Solutions and Their Business Requirements
  Using Feature Interactions.
\newblock In \emph{Proceedings of the 2019 27th ACM Joint Meeting on European
  Software Engineering Conference and Symposium on the Foundations of Software
  Engineering}, ESEC/FSE 2019, 1048–1058. New York, NY, USA: Association for
  Computing Machinery.
\newblock ISBN 9781450355728.

\bibitem[{Galton(1877)}]{galton1877typical}
Galton, F. 1877.
\newblock \emph{Typical laws of heredity}.
\newblock William Clowes and Sons.

\bibitem[{Ghojogh and Crowley(2019)}]{QDA}
Ghojogh, B.; and Crowley, M. 2019.
\newblock Linear and Quadratic Discriminant Analysis: Tutorial.
\newblock arXiv:1906.02590.

\bibitem[{Guan and Loew(2021)}]{DistSepIdx}
Guan, S.; and Loew, M. 2021.
\newblock A distance-based separability measure for internal cluster
  validation.
\newblock \emph{arXiv}.

\bibitem[{Guyon and Elisseeff(2003)}]{sepFeature1}
Guyon, I.; and Elisseeff, A. 2003.
\newblock An Introduction to Variable and Feature Selection.
\newblock \emph{J. Mach. Learn. Res.}, 3(null): 1157–1182.

\bibitem[{H. et~al.(1996)H., C., L., A., and V.}]{SVC}
H., D.; C., B.; L., K.; A., S.; and V., V. 1996.
\newblock Support vector regression machines.
\newblock \emph{Advances in Neural Information Processing Systems 9}.

\bibitem[{Hemphill, Godfrey, and Doddington(1990)}]{ATIS}
Hemphill, C.~T.; Godfrey, J.~J.; and Doddington, G.~R. 1990.
\newblock The ATIS Spoken Language Systems Pilot Corpus: Proceedings of a
  Workshop Held at Hidden Valley, Pennsylvania, June 24-27, 1990.
\newblock {www.kaggle.com/siddhadev/atis-dataset-from-ms-cntk}.

\bibitem[{Keikha, Park, and Croft(2014)}]{webap_data}
Keikha, M.; Park, J.~H.; and Croft, W.~B. 2014.
\newblock Evaluating Answer Passages using Summarization Measures.
\newblock \emph{SIGIR}.
\newblock {https://ciir.cs.umass.edu/downloads/WebAP/}.

\bibitem[{Ledoit and Wolf(2004)}]{ledoit2004well}
Ledoit, O.; and Wolf, M. 2004.
\newblock A well-conditioned estimator for large-dimensional covariance
  matrices.
\newblock \emph{Journal of multivariate analysis}, 88(2): 365--411.

\bibitem[{Lee et~al.(2018)Lee, Lee, Lee, and Shin}]{lee2018simple}
Lee, K.; Lee, K.; Lee, H.; and Shin, J. 2018.
\newblock A simple unified framework for detecting out-of-distribution samples
  and adversarial attacks.
\newblock \emph{Advances in neural information processing systems}, 31.

\bibitem[{Liu et~al.(2019)Liu, Eshghi, Swietojanski, and
  Rieser}]{liu2019benchmarking}
Liu, X.; Eshghi, A.; Swietojanski, P.; and Rieser, V. 2019.
\newblock Benchmarking natural language understanding services for building
  conversational agents.
\newblock \emph{arXiv preprint arXiv:1903.05566}.

\bibitem[{McLachlan(1999)}]{mclachlan1999mahalanobis}
McLachlan, G.~J. 1999.
\newblock Mahalanobis distance.
\newblock \emph{Resonance}, 4(6): 20--26.

\bibitem[{Misra(2018)}]{misra2018news}
Misra, R. 2018.
\newblock News Category Dataset.

\bibitem[{Misra and Grover(2021)}]{misra2021sculpting}
Misra, R.; and Grover, J. 2021.
\newblock \emph{Sculpting Data for ML: The first act of Machine Learning}.
\newblock ISBN 9798585463570.

\bibitem[{Mthembu and Marwala(2008)}]{noteOnSep}
Mthembu, L.; and Marwala, T. 2008.
\newblock A note on the separability index.
\newblock arXiv:0812.1107.

\bibitem[{Navot et~al.(2005)Navot, Gilad-Bachrach, Navot, and
  Tishby}]{setFeature2}
Navot, A.; Gilad-Bachrach, R.; Navot, Y.; and Tishby, N. 2005.
\newblock Is Feature Selection Still Necessary?
\newblock In \emph{International Statistical and Optimization Perspectives
  Workshop "Subspace, Latent Structure and Feature Selection"}, 127--138.
\newblock ISBN 978-3-540-34137-6.

\bibitem[{Peterson(2011)}]{sepIdx}
Peterson, A.~D. 2011.
\newblock \emph{A separability index for clustering and classification problems
  with applications to cluster merging and systematic evaluation of clustering
  algorithms}.
\newblock Ph.D. thesis, Iowa State University.

\bibitem[{Reimers and Gurevych(2019)}]{reimers-2019-sentence-bert}
Reimers, N.; and Gurevych, I. 2019.
\newblock Sentence-BERT: Sentence Embeddings using Siamese BERT-Networks.
\newblock In \emph{Proceedings of the 2019 Conference on Empirical Methods in
  Natural Language Processing}. Association for Computational Linguistics.

\bibitem[{Thornton(2008)}]{ThorntonSepIdx}
Thornton, C. 2008.
\newblock Separability is a Learner’s Best Friend.
\newblock In \emph{Proceedings of the Fourth Neural Computation and Psychology
  Workshop}.
\newblock ISBN 978-3-540-76208-9.

\end{thebibliography}
}

\end{document}